\def\year{2017}\relax
\pdfoutput=1

\documentclass[letterpaper]{article}
\usepackage{icwsm17}
\usepackage{color}
\usepackage{amsmath}
\usepackage{times}
\usepackage{helvet}
\usepackage{courier}
\usepackage{graphicx}
\usepackage{array, booktabs}
\newcolumntype{C}[1]{>{\centering\let\newline\\\arraybackslash\hspace{0pt}}m{#1}}

\usepackage{subfigure}
\usepackage[]{algorithm2e}
\newcommand{\citenoun}[1]{{\citeauthor{#1} \shortcite{#1}}}
\newcommand{\cut}[1]{{}}
\newcommand{\outline}[1]{{}}
\newcommand{\x}{\mathbf{x}}

\newcommand{\z}{\mathbf{z}}

\newcommand{\argmax}{\operatornamewithlimits{arg\,max}}

\newenvironment{packed_item}{
\begin{itemize}
  \setlength{\itemsep}{1pt}
  \setlength{\parskip}{0pt}
  \setlength{\parsep}{0pt}
}{\end{itemize}}

\newenvironment{packed_enum}{
\begin{enumerate}
  \setlength{\itemsep}{1pt}
  \setlength{\parskip}{0pt}
  \setlength{\parsep}{0pt}
}{\end{enumerate}}

\begin{document}
%
\title{Controlling for Unobserved Confounds in Classification \\ Using Correlational Constraints}
\author{
  Virgile Landeiro \and Aron Culotta\\
  Department of Computer Science\\
  Illinois Institute of Technology\\
  Chicago, IL 60616\\
  vlandeir@hawk.iit.edu, aculotta@iit.edu\\
}
\maketitle
\begin{abstract}
    \begin{quote}
      As statistical classifiers become integrated into real-world  applications, it is important to consider not only their accuracy but also their robustness to changes in the data distribution. In this paper, we consider the case where there is an unobserved confounding variable $z$ that influences both the features $\x$ and the class variable $y$. When the influence of $z$ changes from training to testing data, we find that the classifier accuracy can degrade rapidly. In our approach, we assume that we can predict the value of $z$ at training time with some error. The prediction for $z$ is then fed to Pearl's back-door adjustment to build our model. Because of the attenuation bias caused by measurement error in $z$, standard approaches to controlling for $z$ are ineffective. In response, we propose a method to properly control for the influence of $z$ by first estimating its relationship with the class variable $y$, then updating predictions for $z$ to match that estimated relationship. By adjusting the influence of $z$, we show that we can build a model that exceeds competing baselines on accuracy as well as on robustness over a range of confounding relationships.
    \end{quote}
\end{abstract}

\section{Introduction}

\outline{
- classifiers are used to make important decisions about people (sentencing, credit scores, job offers)
- classifiers are often treated as black boxes
- Many biases can exist in the data. In this paper, we're concerned about confounders (define)
- this is particularly a problem when
  - many features are used (hard to identify and diagnose bias)
  - data changes over time ($P_{train}(x,y) != P_{test}(x,y)$
- Emerging studies on fairness in machine learning (cite them)
- Recently, Virgile introduce a method based on BDA
- But, assumes Z is observed at training time. Very limiting assumption, since it is hard to collect data labeled across many dimensions (e.g., all SES)
- Simple alternative is to build a classifier for Z, and use it to label Y's training data
- But, this has big problems:
  - attenuation bias: reduces BDA's ability to adjust for confounder
- We propose two solution:
  - Just use confidence as threshold
    - but this wastes data
  - Estimate error rate of classifier, estimate true correlation in Y training data, optimize directly
- We show that correlation optimization improves robustness, and also improves accuracy of Z
}

Statistical classifiers have become widely used to inform important decisions such as  whether to approve a loan~\cite{hand1997statistical}, hire a job candidate~\cite{miller2015can}, or release a criminal defendant on bond~\cite{monahan2016risk}.  Given the significant real-world consequences of such decisions, it is critical that we can identify and remove sources of systematic bias in classification algorithms. For example, some evidence suggests that existing criminal recidivism models may be racially biased~\cite{angwin2016machine}.

One important type of classifier bias arises from {\it confounding variables}. A confounder $z$ is a variable that is correlated both with the input variables (or features) $\x$ and the target variable (or label) $y$ of a classifier. When $z$ is not included in the model, the true relationship between $\x$ and $y$ can be improperly estimated; in the social sciences -- originally in econometrics -- this is called omitted variable bias. While omitted variable bias is a core focus of social science~\cite{king1994designing}, it has received much less attention in machine learning communities, where prediction accuracy is the main concern. Confounding variables can be particularly problematic in high-dimensional settings, such as text classification, where models may contain thousands or millions of parameters, making manual inspection of models impractical. The common use of text classification in computational social science applications~\cite{lazer2009life} further adds to the urgency of the problem.

Several studies with interests in public health focused on tracking the influenza rates in the USA by using Twitter as a sensor \cite{paul2011you}. These studies demonstrated that machine learning offers more accurate, inexpensive, and fast tracking methods than what is currently used by the CDC. \citeauthor{de2013predicting} built models to predict postpartum changes in emotion and behavior using Twitter data and managed to identify mothers who will change significantly following childbirth with an accuracy of 71\% using observations about their prenatal behavior~\cite{de2013predicting}. In a more recent study, \citeauthor{koratana2016studying} collected Yik Yak data -- an anonymous social network popular among students -- to study anonymous health issues and substance use on college campuses~\cite{koratana2016studying}. The results of these studies are encouraging for the field of computational social science but only a few of them are taking into account the effect of possible confounders. A growing body of work tries to mitigate the effect of observed confounding variables using causal inference techniques. For instance, \citeauthor{cunha2017warm} use a matching approach for causal inference to estimate the effect of online support on weight loss using data from Reddit, and \citeauthor{de2016discovering} leverage propensity score matching to detect users that transition from posting about mental health concerns to posting about suicidal ideation on Reddit.
In this paper, we wish to provide methods for researchers in computational social sciences to conduct observational studies while controlling for confounding variables even though these might not be directly observed.  

In recent work~\cite{landeiro2016robust}, a text classification algorithm was proposed based on Pearl's back-door adjustment \cite{pearl2003causality} as a framework for prediction that controls for an observed confounding variable. It was found that this approach results in classifiers that are significantly more robust to shifts in the relationship between confounder $z$ and class label $y$. However, an important limitation of this prior work is that it assumes that a training set is available in which every instance is annotated for both class label $y$ and confounder $z$. This is problematic because there are many confounders we may want to control for (e.g., income, age, gender, race/ethnicity) that are often rarely available and difficult for humans to label, particularly in addition to the primary label $y$.

A natural solution is to build statistical classifiers for confounders $z$, and use the predicted values of $z$ to control for these confounders. However, the measurement error of $z$ introduces {\it attenuation bias}~\cite{chesher1991effect} in the back-door adjustment, resulting in classifiers that are still confounded by $z$.

In this paper, we present a classification algorithm based on Pearl's back-door adjustment to control for an {\it unobserved} confounding variable. Our approach assumes we have a preliminary classifier that can predict the value of the confounder $z$, and that we have an estimate of the  error rate of this $z$-classifier. We offer two methods to adjust for the mislabeled $z$ to improve the effectiveness of back-door adjustment. A straightforward approach is to remove training instances for which the confidence of the predicted label for $z$ is too low. While we do find this approach can reduce attenuation bias, it must discard many training examples, degrading the $y$-classifier. Our second approach instead uses the error rate of the $z$-classifier to estimate the correlation between $y$ and $z$ in the training set. The assignment to $z$ is then optimized to match this estimated correlation, while also maximizing classification accuracy. We compare our methods on two real-world text classification tasks: predicting the location of a Twitter user and predicting if a Twitter user is smoking or not. Both prediction tasks are using users' tweets as input data and are confounded by gender. The resulting model exhibits significant improvements in both accuracy and robustness, with some settings producing similar results as fully-observed back-door adjustment.

\cut{
Our second approach is to leverage the knowledge we have of the preliminary study. By estimating the true correlation $r(y,z)$ and then updating $z$ predictions to estimate that true influence, we show that we can build a model approaching the performance of a back-door adjustment model controlling for an observed confounding variable both in accuracy and robustness.

First, we show by example that back-door adjustment performance and robustness are highly dependent of the quality of the preliminary study.
With more data made openly available every day, statistical analysis tools such as classification models are used by researchers across fields such as social science (), political science (), or medical research ().
To ensure the validity of these studies - with conceivably meaningful repercussion - it is critical to not only consider the accuracy of a classification model but also its robustness to changes in the data distribution.

We evaluate our approaches on a text classification task: predicting the location of a Twitter user given his/her tweets (confounded by gender).
}

\section{Related Work}
\label{sec:related-work}

In the machine learning field, selection bias has received some attention~\cite{zadrozny2004learning,bareinboim2014recovering}. It arises when the population of a study is not selected randomly. Instead, some users are more inclined to be selected for the study than others, making it more difficult to draw conclusions from the general population. If we denote $S$ whether or not an element of the population is selected, there is presence of selection bias when $p(S=1|X,Y) \neq p(S=1)$. Dataset shift~\cite{quionero2009dataset} is a similar issue that appears when the joint distribution of features and labels changes between the training dataset and the testing dataset (i.e. $p_{tr}(X,Y) \neq p_{te}(X,Y)$). Covariate shift~\cite{bickel2009discriminative,sugiyama2007covariate} is a specific case of dataset shift in which only the inputs distribution is different from training to testing (i.e. $p_{tr}(X) \neq p_{te}(X)$). Similarly, when the underlying target distribution $p(Y)$ changes over time, either in a sudden way or gradually, then this is called concept drift~\cite{tsymbal2004problem,widmer1996learning}. Recent work has studied ``fairness'' in machine learning~\cite{zemel2013learning,hajian2013methodology} as well as attempted to remove features that introduce bias~\cite{pedreshi2008discrimination,fukuchi2013prediction}. \citenoun{kuroki2014measurement} propose an extension of back-door adjustment to deal with measurement error in the confounder, but it does not scale well when $\x$ is high dimensional, as in our setting of text classification. 
 
Although all these types of biases are important to conduct a valid observational study, in this paper we direct our attention to the problem of learning under confounding bias shift. In other words, we aim to build a classifier that is robust to changes in the relation between the target variable $Y$ of a classifier and an external confounding variable $Z$. \citenoun{landeiro2016robust} use back-door adjustment for text classification, but assume confounders are observed at training time. This paper introduces methods to enable back-door adjustment to work effectively when confounders are unobserved and when the features are high dimensional.

\section{Methods}

In this section, we first review prior work using back-door adjustment to control for observed confounders in text classification. We then introduce two methods for applying back-door adjustments when the confounder is unobserved at training time and must instead be predicted by a separate classifier.

\subsection{Adjusting for {\it observed} confounders}

Suppose one wishes to estimate the causal effect of a variable $\x$ on a variable $y$ when a randomized controlled trial is not possible. If a sufficient set of confounding variables $z$ is available, one can use the back-door adjustment equation as follows:

\begin{equation}
  p(y|do(\x)) = \sum_{z}{p(y|\x,z)\times p(z)}
 \label{eq:ba}
\end{equation}

The {\it back-door criterion}~\cite{pearl2003causality} is a graphical test that determines whether $z$ is a sufficient set of variables to estimate the causal effect. This criterion requires that no node in $z$ is a descendant of $\x$ and that $z$ blocks every path between $\x$ and $y$ that contains an arrow pointing to $\x$. Notice $p(y|\x) \neq p(y|\text{do}(\x))$: this do-notation is used in causal inference to indicate that an intervention has been made on $\x$. \cut{Figure~\ref{fig:model} displays the directed graphical model for the back-door adjustment.} Omitting the predicted confounder $z'$, it depicts a standard discriminative approach to classification, e.g., modeling $p(y|\x)$ with a logistic regression classifier conditioned on the observed term vector $\x$. We assume that the confounder $z'$ influences both the term vector through $p(x|z)$ as well as the target label through $p(y|z')$. The structure of this model ensures that $z'$ meets the back-door criterion for adjustment.

Back-door adjustment was originally introduced for causal inference problems --- i.e., to estimate the causal effect of performing action $\x$ on outcome $y$. Recently, \citenoun{landeiro2016robust} have shown that back-door adjustment can also be used to improve classification robustness. By controlling for a confounder $z$, the resulting classifier is robust to changes in the relationship between $z$ and $y$.

From the perspective of standard supervised classification, the approach works as follows: Assume we are given a training set $D=\{(\x_i, y_i)\}$. If we suspect that a classifier trained on $D$ is confounded by some additional variable $z$, we augment the training set by including $z$ as a feature for each instance: $D'=\{(\x_i, y_i, z_i)\}$. We then fit a classifier on $D'$, and at testing time apply Equation \ref{eq:ba} to classify new examples --- $p(y|\x)  = \sum_z p(y|\x, z)p(z)$ --- where $p(z)$ is simply computed from the observed frequencies of $z$ in $D'$. By controlling for the effect of $z$, the resulting classifier is robust to the case where $p(y|z)$ changes from training to testing data.

In the experiments below, we consider the problem of predicting a user's location $y$ based on the text of their tweets $\x$, confounded by the user's gender $z$. That is, in the training data, there exists a correlation between gender and location, but we want the classifier to ignore that correlation. When the above procedure is applied to a logistic regression classifier, the result is that the magnitudes of coefficients for terms that correlate with gender are greatly reduced, thereby minimizing the effect of gender on the classifier's predictions.

\cut{
If we are given a training set $D = \{(\x_i, y_i, z_i)\}_{i=1}^n$, where each instance consists of a vector of features $\x$, a label $y$, and a confounding variable $z$, we can predict the label $y_j$ for some new instance $\x_j$, while controlling for an unobserved confounder $z_j$. That is, we assume $z$ is observed at training time, but not at testing time.

When $z$ is observed, Pearl's back-door adjustment  can be used as a prediction framework to find the underlying relationship between $\x$ and $y$.
}

\cut{
\begin{figure}[t]
  \centering
  \includegraphics[width=0.3\textwidth]{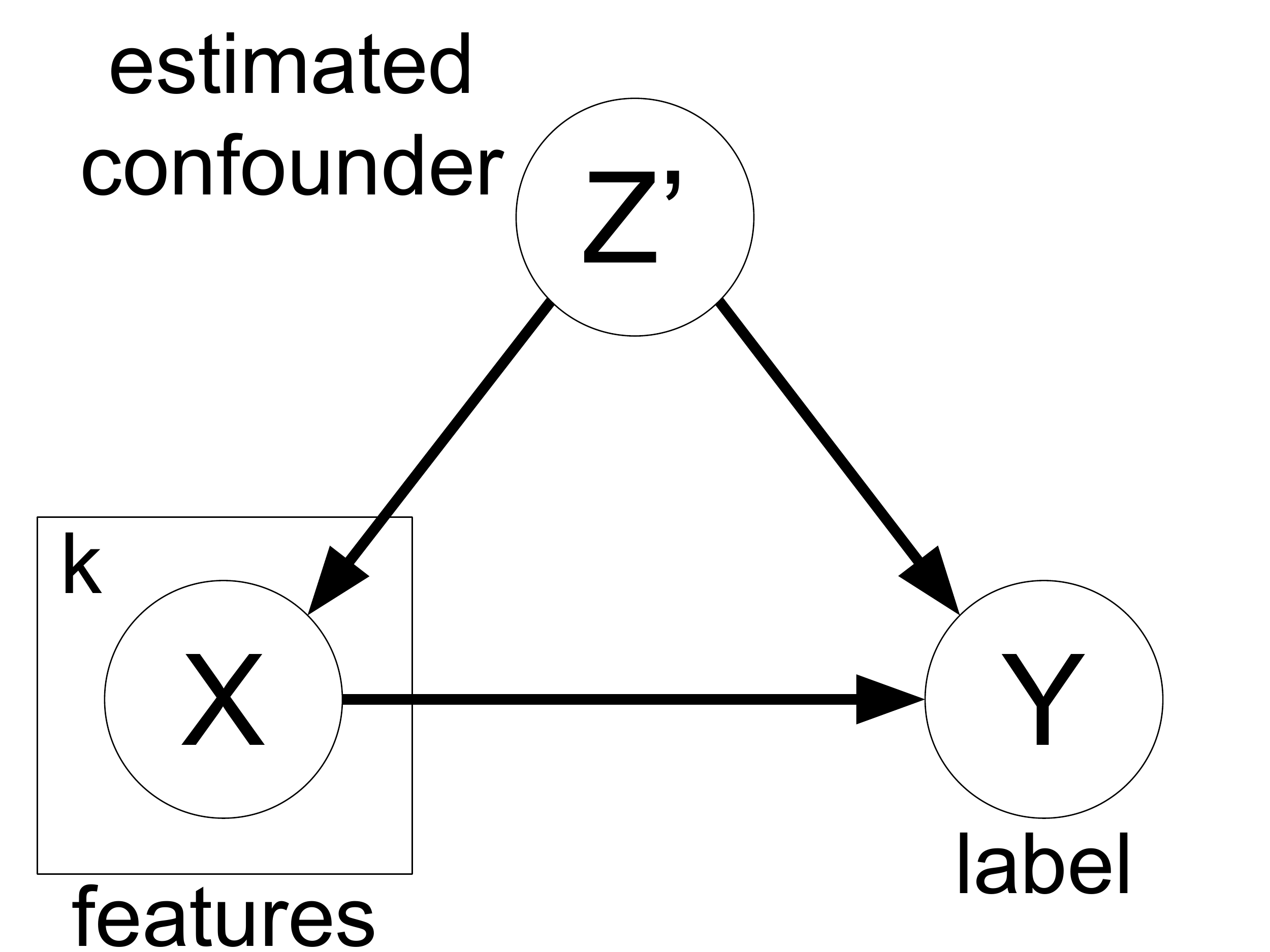}
  \caption{Directed graphical model depicting a confounding variable $z'$ influencing both observed features $\x$ and class variable $y$.}\label{fig:model}
\end{figure}
}

\subsection{Adjusting for {\it unobserved} confounders}

In the previous approach, it was assumed that we had access to a training set $D=\{(\x_i, y_i, z_i)\}$; that is, each instance is annotated both for the label $y$ and confounder $z$. This is a burdensome assumption, given that ultimately we will need to control for many possible confounders (e.g., gender, race/ethnicity, age, etc.). Because many of these confounders are unobserved and/or difficult to obtain, it is necessary to develop adjustment methods that can handle noise in the assignment to $z$ in the training data.

Our approach assumes we have an (imperfect) classifier for $z$, trained on a secondary training set $D_z=\{(\x_i, z_i)\}$  --- we call this the {\it preliminary study}, with the resulting {\it preliminary classifier} $p(z|\x)$. This is combined with the dataset $D_y=\{(\x_i, y_i)\}$, used to train the primary classifier $p(y|\x)$. The advantage of allowing for separate training sets $D_y$ and $D_z$ is that it is often easier to annotate $z$ variables for some users than others; for example, \citenoun{pennacchiotti2011machine} build training data for ethnicity classification by searching for online users that explicitly state their ethnicity in their user profiles.

After training on $D_z$, the preliminary classifier is applied to $D_y$ to augment it with predicted annotations for confounder $z$: $D = \{(\x_i, y_i, z'_i)\}_{i=1}^n$, where $z'_i$ denotes the predicted value of $z_i$. A tempting approach is to simply apply back-door adjustment as usual to this dataset, ignoring the noise introduced by $z'$. However, the resulting classifier will no longer properly control for the confounder $z$ for at least two related reasons:
\begin{packed_enum}
  \item The observed correlation between $y$ and $z'$ in the training data will underestimate the actual correlation (i.e., $|r(y,z')| < |r(y, z)|$). This {\it attenuation bias} reduces the coefficients for the $z$ features, which in turn prevents back-door adjustment from reducing the coefficients of features in $\x$ that correlate with $z$.
  \item Similarly, because some training instances have mislabeled annotations for $z$, it is more difficult to detect which features in $\x$ correlate with $z$, thereby preventing back-door adjustment from reducing those coefficients.
\end{packed_enum}

\cut{With perfect annotations of $z$ (i.e.observed confounder), back-door adjustment leverages undertraining and puts more weights on the features encoding for the confounding variable, leading to reduce the coefficients of confounder-related features. }

\cut{Although it has been shown \cut{by \citenoun{landeiro2016robust}} that back-door adjustment provides a robust and accurate prediction when $z$ is observed at training time, here we focus on the case where $z$ is unobserved. Instead, we assume $z$ can be predicted with some error at training time using a model trained beforehand on an unbiased dataset $D_{pre} = \{w_k, z_k\}_{k=1}^m$. The output of this model is noted $z'$ as it is an approximation of $z$.}

\begin{figure}[t]
  \centering
  \includegraphics[width=0.4\textwidth]{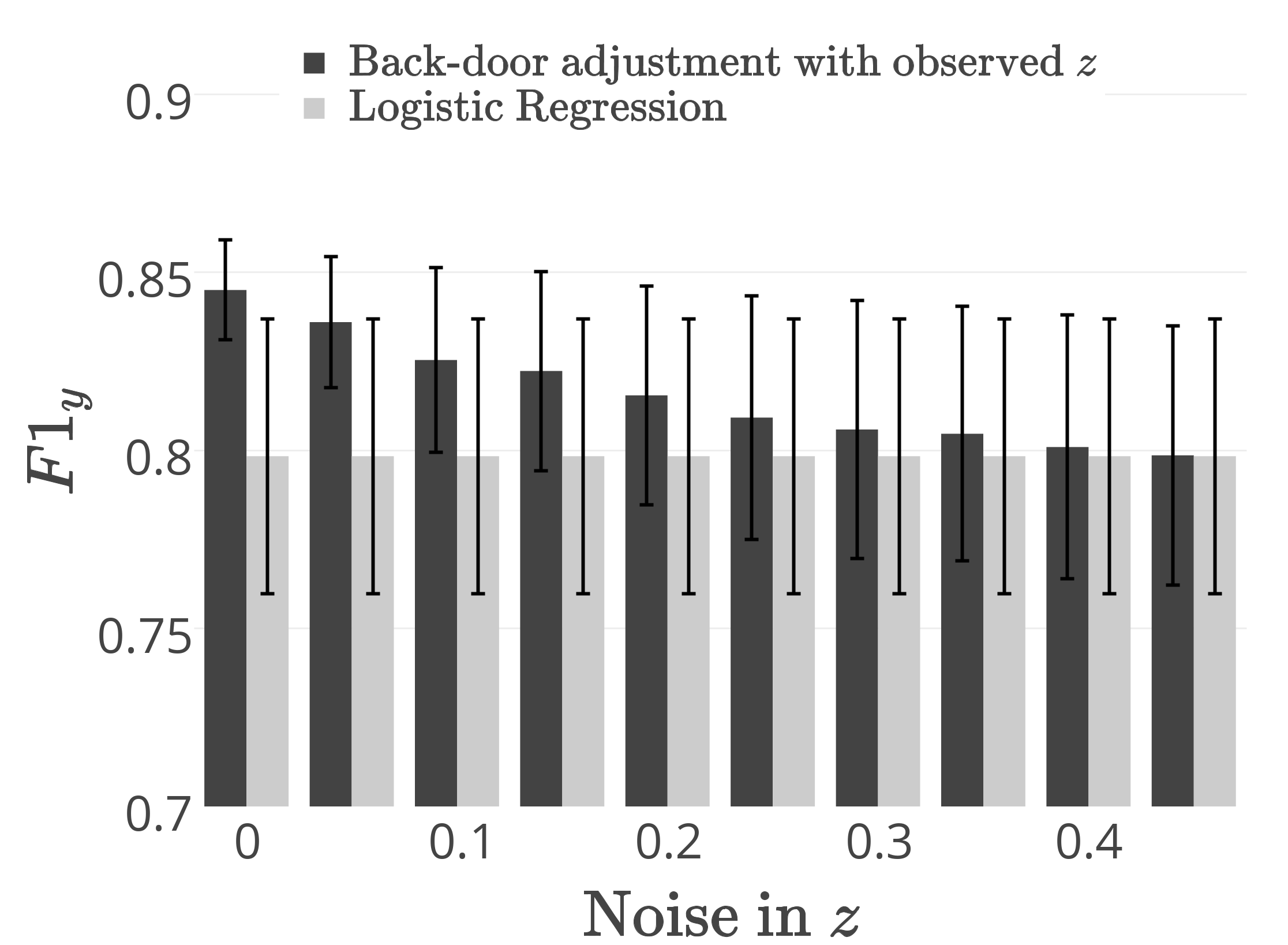}
  \caption{As measurement error in confounder $z$ increases, the effectiveness of back-door adjustment decreases.}\label{fig:broken-ba-vs-lr}
\end{figure}

\cut{Therefore, using back-door adjustment with the raw predictions of the preliminary classifier at training time is not as robust as back-door adjustment when the confounder is observed at training time.}

To verify this claim, we conduct an experiment in which we observe $z$ but we inject increasing amounts of noise in $z$ (e.g., with probability $p$, change the assignment to $z_i$ to be incorrect). In other words, we synthetically decrease the quality of our observations of $z$ and we observe how that influences the performance of back-door adjustment. We then measure how the accuracy of the primary classifier for $y$ varies on a testing set in which the influence of $z$ is decreased (i.e., $z$ correlates strongly with $y$ in the training set, but only weakly in the testing set). These experiments will be discussed in more detail in Section~\ref{sec:experiments}.

We can see in Figure \ref{fig:broken-ba-vs-lr} that the F1 score quickly decreases as we add more noise to the confounding variable annotations, indicating the need for new methods to adjust for unobserved confounders. Notice that when noise is 0, back-door adjustment greatly improves F1 (from .79 F1 with no adjustment to .85 F1), demonstrating the effectiveness of this approach when the confounder is observed at training time. In the following two sections, we propose two methods to fix these issues. \cut{Both aim to get a better estimate of $r(y,z)$ and reduce the amount of mislabeled annotations in $z'$ though the approach the problems differently.}

\cut{
 The heatmap shows the F1 performance of the back-door adjustment given the noise in the confounding variable (i.e.$p_flip$) on the y-axis and the correlation difference between the training and the testing set on the x-axis. In other words, the x-axis is the difference between the correlation between $y$ and $z$ in the training dataset and the correlation between $y$ and $z$ in the testing dataset. In this figure, the F1 score of the $y$ classifier is displayed using a color scale and the robustness to the changes in the data is shown by the error bars on the left scatter plot. Because these error bars show the standard deviation of the F1 score given a fixed noise value, a more robust classifier will display smaller error bars.
}

\begin{table}[t]
\centering
\begin{tabular}{lrrrrrr}
  \toprule
  \textbf{Noise} & 0.00 & 0.05 & 0.10 & 0.15 & 0.20  \\
  \textbf{F1 std dev} & 0.028 & 0.037 & 0.052 & 0.056 & 0.062\\
\bottomrule
\end{tabular}
\caption{Evolution of the standard deviation of F1 scores in back-door adjustment for a given noise in $z$. The lower the standard deviation, the more robust the model.}
\end{table}

\subsubsection{Thresholding on confidence of $z$ predictions}

Our first approach is fairly simple; its objective is to directly reduce the number of mislabeled annotations in $z'$. Our preliminary model produces the value $z'_i$ (the prediction of the true confounder $z_i$) as well as $p(z_i=z'_i|\x_i)$ (the confidence of the prediction; i.e., the posterior distribution over $z$). We use these posteriors to remove predictions with low confidence. By setting a threshold $\epsilon \in [0.5, 1]$, we filter the original dataset $D=\{\x_i, y_i, z'_i\}$ by keeping an instance $i$ only if it satisfies $p(z_i=z'_i|\x_i) \ge \epsilon$.

For well-calibrated classifiers like logistic regression, we expect to remove mostly mislabeled data points by thresholding at $\epsilon$. Making $\epsilon$ vary between $0.5$ and $1$ allows us to modify the output of the preliminary study in order to obtain a sub-dataset with as many points correctly labeled as possible. Moreover, when the error of our preliminary classifier is symmetric, this process will also move the estimated correlation $r(y,z')$ towards the true correlation $r(y,z)$.

With this smaller set of training instances, we run back-door adjustment without modification. However, one important drawback of this method is that we remove instances from our training dataset. Depending on the quality of the preliminary classifier and the setting of $\epsilon$, only a small fraction of training instances may potentially remain. Thus, in the next section we consider an alternative approach that does not require discarding training instances.

\subsubsection{Correlation matching}

While the above approach aims to reduce errors in $z'$, and as a side effect improves the estimate of $r(y,z)$, in this section we propose an approach that directly tries to improve the estimate of $r(y,z)$ while also reducing errors in $z$. Let $r' = r(y, z')$ be the observed correlation between $y$ and $z'$, and let $r=r(y, z)$ be the true (unobservable) correlation between $y$ and $z$ in the training data for $y$, $D=\{\x_i, y_i, z_i'\}$. Our proposed approach builds on the insight of \citenoun{francis1999high}, who show that $r'$ can be estimated from $r$ using the variances of $y$ and $z$ as well as the variances of the errors in $y$ and $z$:
\begin{equation}
  r' = \sqrt{\frac{1} { ( 1 + \frac{V_{ey}}{V_y} ) ( 1 + \frac{V_{ez}}{V_z} )}} \times r
\end{equation}
where $V_z$ is the variance of $z$, and $V_{ez}$ is the variance of error on $z$, and analogously for $V_y$, $V_{ey}$. Since in our setting $y$ is observed, we can set $V_{ey} = 0$ and solve for $r$:
\begin{align}
    r' & = \sqrt{\frac{1} {1 + \frac{V_{ez}}{V_z}}} \times r \\
    \Rightarrow r & = r' \times \sqrt{ 1+\frac{V_{ez}}{V_z} } \label{eq:r_est}
\end{align}
Thus, the factor by which $r'$ underestimates $r$ is proportional to the ratio of the variance of the error in $z$ to the variance of $z$.

We can estimate the terms $V_z$ and $V_{ez}$ using cross-validation on the preliminary training data $D_z = \{(\x_i, z_i)\}$. Let $z'_i$ be the value predicted by the preliminary classifier on instance $\x_i \in D_z$, where $i$ is in the testing fold of one cross-validation split of the data. Let $e^z_i=|z_i - z'_i|$ be the absolute error of $z$ on instance $i$. Then, we can first compute the mean absolute error of $z'_i$ as
$\mu_{ez} = \frac{1}{|D_z|} \sum_{z_i \in D_z}e_i^z$. The estimated variance of the errors in $z$ is then:
\begin{equation}
\hat{V}_{ez} = \frac{1}{|D_z|}\sum_{z \in D_z}(e^z_i - \mu_{ez})^2
\label{eq:err_var}
\end{equation}

Since this variance in the error of $z$ in turn affects the observed variance of $z$, we can then estimate
\begin{equation}
\hat{V}_z = V_{z'} - \hat{V}_{ez}
\label{eq:z_var}
\end{equation}
where $V_{z'}$ is the variance of predictions $z'$ in the target training data $D$.

Plugging the estimates of Equations \ref{eq:err_var} and \ref{eq:z_var} into Equation \ref{eq:r_est} enables us to estimate the true correlation between $y$ and $z$ in the target training data $D$. We will refer to this estimated correlation as $\hat{r}$.

\cut{If we assume that the errors in $z'$ are unbiased, then we can assume $\bar{z}' = \bar{z}$.
$\hat{V}_z = V_{z'} - V_{ez}$}

As an example, consider a dataset $D = \{(\x_i, y_i, z'_i)\}$. The original correlation $r(y, z') \equiv r'$ may be .5, but the true correlation $r(y,z) \equiv r$ may be .8. Depending on the variances of $z$ and its error, the estimated correlation may be $\hat{r}=.75$. The next step in the procedure is to optimize the assignment to $z'$ to minimize the difference $|r' - \hat{r}|$. That is, we use $\hat{r}$ as a soft constraint, and attempt to match that constraint by changing the assignments to $z'$.

Let $\mathbf{Z}$ be the set of all possible assignments to $z$ in the training set $D$ (i.e., if $z$ is a binary variable and $|D|=n$, then $|\mathbf{Z}|= 2^n$). Let $\z^j = \{z_1^j \ldots z_n^j\} \in \mathbf{Z}$ be a vector of assignments to $z$, and let $r'(\z^j)$ indicate the correlation  $r(\z^j,y)$. Then our objective is to choose an assignment from $\mathbf{Z}$ to minimize $r'(\z^j) - \hat{r}$, while still maximizing the probability of that assignment according to the preliminary classifier for $z$. We can write this objective as follows:
\begin{equation}
\z^* \leftarrow   \argmax_{\z^j \in \mathbf{Z}} \left( \frac{1}{n}\sum_{z_i^j \in \z^j} p(z_i=z_i^j|\x_i)\right) - |\hat{r} - r'(\z^j)|
\end{equation}
Thus, we search for an optimal assignment $\z^*$ that maximizes the average posterior of the predicted $z$ value, while minimizing the difference between the estimated correlation $\hat{r}$ and the observed correlation $r'(\z^j)$.

This optimization problem can be approached in several ways. We implement a greedy hill-climbing algorithm that iterates through the values in $z'$ sorted by confidence and flips the value if it reduces $|r-r'|$. The steps are as follows:
\begin{packed_enum}
\item Initialize $\z^j$ to the most probable assignment according to $p(z|\x)$.
\item Initialize $\mathcal{I}$ to be all instances sorted in descending order of confidence $p(z|\x)$.
\item While $|\hat{r} - r'(\z^j)|$ is decreasing:
  \begin{packed_enum}
  \item Pop the next instance $(\x_i, z_i^j, y_i)$ from $\mathcal{I}$
  \item If flipping the label $z_i^j$ reduces the error  $|\hat{r} - r'(\z^j)|$, do so. Else, skip to the next instance.
  \end{packed_enum}
\item Return the final $\z^j$.
\end{packed_enum}

For example, consider the case where $r'(\z^j) < \hat{r}$. If the instance popped in step 3(a) has labels ($y_i=1$, $z_i'=0$), then we know that flipping $z_i$ to 1 would increase the correlation between $y$ and $z'$. By considering flips in descending order of $p(z|\x)$, we ensure that we first flip assignments that are likely to be incorrect. In the experiments below, we find that this approach often converges after a relatively small number of flips.

The advantages of this approach are that it not only produces assignments to $z$ that better align with the expected correlation $\hat{r}$, but it also results in more accurate assignments to $z$. The latter is possible because we are using prior knowledge about the relationship between $z$ and $y$ to assign values of $z$ when the classifier is uncertain. As with the thresholding approach of the previous section, once the new assignments to $z$ are found, back-door adjustment is run without modification.

\section{Experiments}
\label{sec:experiments}

\cut{Using a dataset extracted using Twitter's API} We conducted text classification experiments in which the relationship between the confounder $z$ and the class variable $y$ varies between the training and testing set. We consider the scenario in which we directly control the discrepancy between training and testing. Thus, we can determine how well a confounder has been controlled by measuring how robust the method performs across a range of discrepancy levels.

To sample train/test sets with different $p(y|z)$ distributions, we assume we have labeled datasets $D_{train}$, $D_{test}$, with elements $\{(\x_i, y_i, z_i)\}$, where $y_i$ and $z_i$ are binary variables. We introduce a bias parameter $p(y=1|z=1) = b$; by definition, $p(y=0|z=1)=1-b$. For each experiment, we sample without replacement from each set $D'_{train} \subseteq D_{train}$, $D'_{test} \subseteq D_{test}$. To simulate a change in $p(y|z)$, we use different bias terms for training and testing, $b_{train}$, $b_{test}$.
We thus sample according to the following constraints: $p_{train}(y=1|z=1) = b_{train}$, $p_{test}(y=1|z=1) = b_{test}$, $p_{train}(Y) = p_{test}(Y)$, and $p_{train}(Z) = p_{test}(Z)$.

The last two constraints are to isolate the effect of changes to $p(y|z)$. Thus, we fix $p(y)$ and $p(z)$, but vary $p(y|z)$ from training to testing data. We emphasize that we do not alter any of the actual labels in the data; we merely sample instances to meet these constraints. In the rest of the paper, we note $r_{train}(y,z)$ (respectively $r_{test}(y,z)$) the correlation between $y$ and $z$ in the training set (resp. testing set). We also denote $\delta_{yz} = r_{train}(y,z) - r_{test}(y,z)$.

\subsection{Datasets}

\subsubsection{Location / Gender}

For our first dataset, we use the data from \citenoun{landeiro2016robust}, where the task is to predict the location of a Twitter user from their messages, with gender as a potential confounder. Thus, $\x$ is a term vector, $y$ is location, and $z$ is gender. The data contain geolocated tweets from New York City (NYC) and Los Angeles (LA). There are 246,930 tweets for NYC and 218,945 for LA over a four-day period (June 15th to June 18th, 2015). \cut{We attempt to filter bots, celebrities, and marketing accounts by removing users with fewer than 10 followers or friends, more than 1,000 followers or friends, or more than 5,000 posts.} Gender labels are derived by cross-referencing the user's name (from the profile) with U.S. Census name data, removing ambiguous names. For each user, we have up to the most recent 3,200 tweets, which we represent each as a single binary unigram vector per user, using standard tokenization. Finally, we subsample this collection and keep the tweets from 6,000 users such that gender and location are uniformly distributed over the users.

\cut{For this paper, we predict  location with  gender as the confounding
variable. Thus, we let $y_i=1$ indicate NYC and $z_i=1$ indicate Male. Due to how we
build this dataset, the data is evenly distributed across the four possible
$y$/$z$ pairs. We refer to this as the {\bf Twitter} dataset.}

\subsubsection{Smoker / Gender}

In our second dataset, the task is to predict if a Twitter user is a smoker or not, with gender as a potential confounder. We start from approx. 3M tweets collected in January and February 2014 using cigarettes related keywords. We randomly pick 40K tweets for which we can identify the user's gender using the Twitter screen name and the U.S. Census name data. We then manually annotate 4.5K of these tweets on whether they show that a user is a smoker (yes) or a non-smoker (no) while discarding uncertain tweets (unknown). We use this data to train a classifier (F1 score = 0.84) to label the remaining 35.5K tweets on the smoker dimension. In order to avoid mislabeled tweets as much as possible, we only keep predictions with a confidence of at least 95\%, yielding an additional 5.5K automatically labeled tweets. These 10K (4.5K manually annotated + 5.5K automatically annotated) tweets have been written by 9K unique users. For each of these users, we collect the most recent tweets (up to 200). Because some users set their profile to be private or because some users that existed in early 2014 have now deleted their account, we obtain at least 20 tweets for 4.6K users. Then we collect all the cigarettes related tweets published by a user in the first two months of 2014 and add them to our dataset. Finally, we balance the dataset on both annotated dimensions by removing users and eventually obtain a dataset of 4084 users.

\begin{figure}[!ht]
  \centering
  \setcounter{subfigure}{0}
  \subfigure[Effect of $\epsilon$ thresholding on $F1_z$ and distance to true correlation.\label{fig:epsilon_thresholding_results}]{\includegraphics[width=0.48\textwidth]{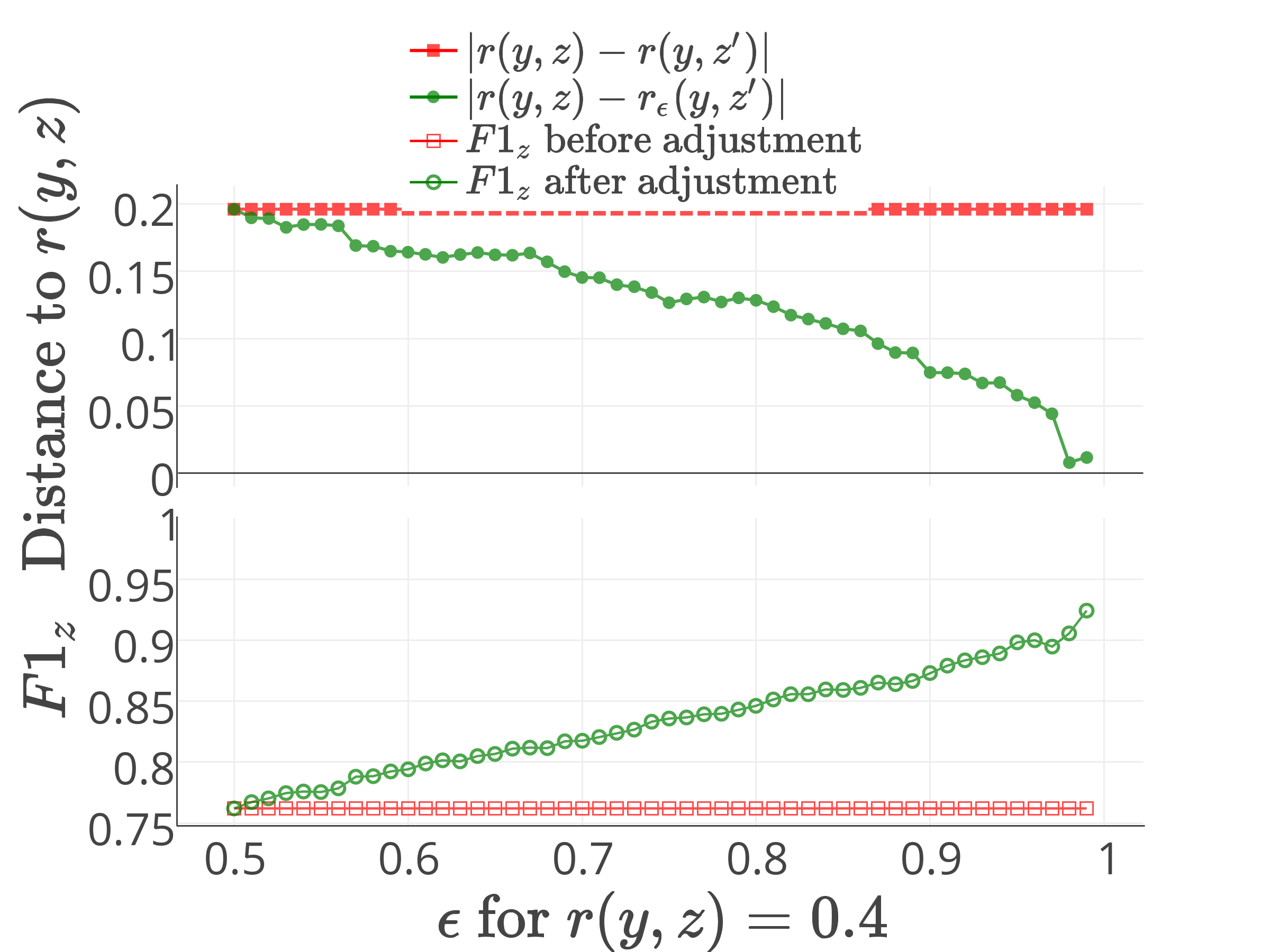}}
  \subfigure[Effect of correlation matching on $F1_z$ and distance to true correlation.\label{fig:corr_matching_results}]{\includegraphics[width=0.48\textwidth]{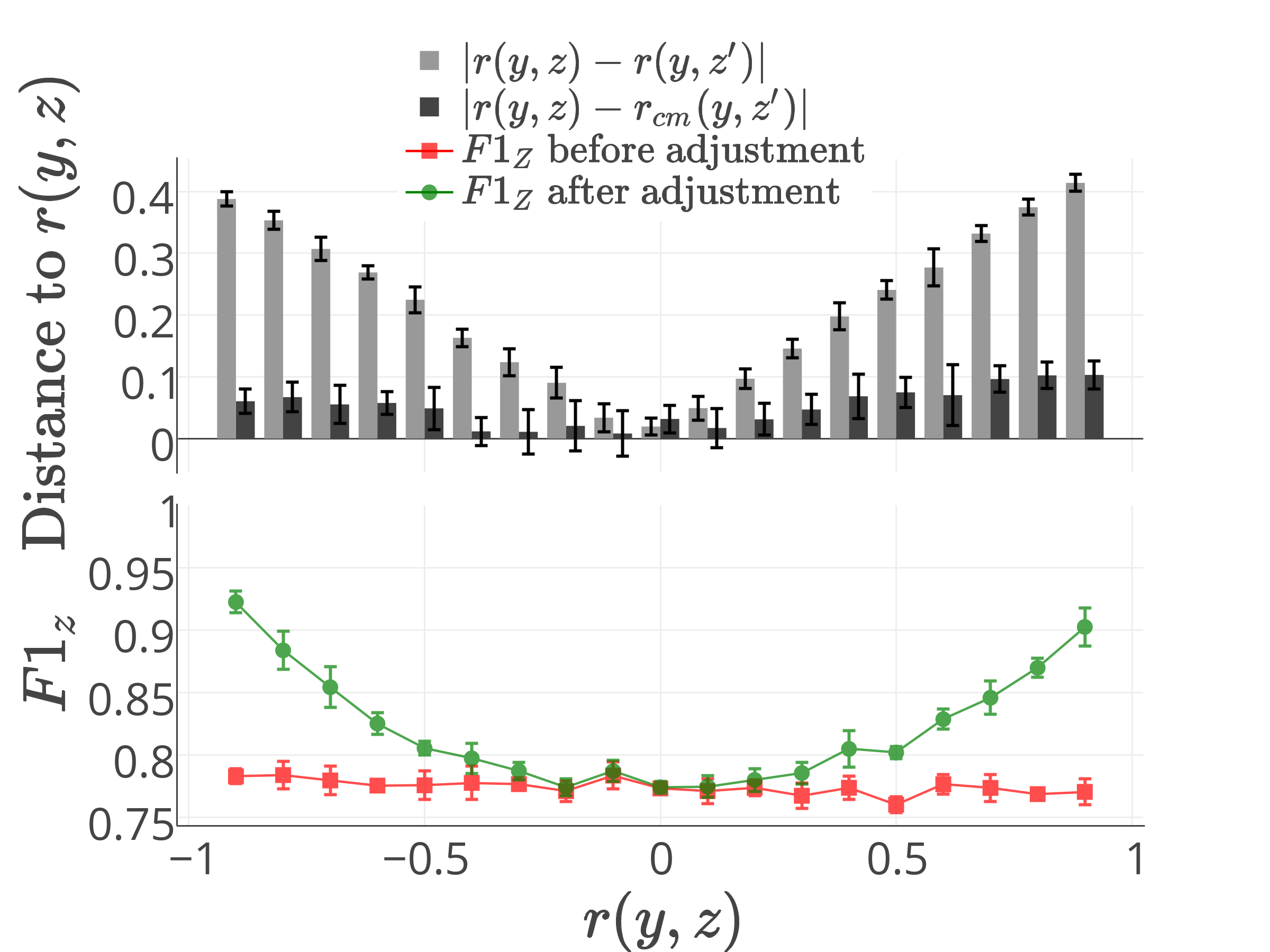}}
  \caption{Effect of correlation adjustment methods.}
\end{figure}

\section{Results}
\label{sec:results}

We use the following notations to describe the results below:
\begin{packed_item}
  \item $\delta_{yz} = r_{train}(y,z) - r_{test}(y,z)$ is the discrepancy between the correlation of $y$ and $z$ in training versus testing.
  \item $r(y,z)$ (respectively $r(y,z')$) is the true (resp. observed) correlation between $y$ and $z$.
  \item $r(y,z_\epsilon')$ (respectively $r(y,z_{cm}')$) is $r(y,z')$ after it has been adjusting using the $\epsilon$ thresholding method (resp. the correlation matching method).
  \item $F1_z$ (respectively $F1_y$) is the F1 score for a $z$ (resp. $y$) classifier, i.e. for the preliminary (resp. main) study.
\end{packed_item}

\cut{
In this section, we first show the advantages and limitations of our two methods to solve the issues caused by the mislabeled elements in $z$, and how they influence the performance of our preliminary study. Then, we will see how these adjustments affect the end results of our $y$ classifier when the association between the target variable and the output variable is different in the training and testings sets.
}

\subsection{Effects of correlation adjustments on $F1_z$}

\noindent For this first part of our results, we obtain quasi-identical outcomes for both datasets. Therefore, we only present the results from the location/gender dataset.

\noindent\textbf{$\mathbf{\epsilon}$ thresholding method:} We make $\epsilon$ vary between $0.5$ and $0.95$ and observe how this reduces the difference between $r(y,z'_{\epsilon})$ and $r(y,z)$. Figure~\ref{fig:epsilon_thresholding_results} shows the result of one setting where $r(y,z) = 0.4$. The figure demonstrates that by increasing $\epsilon$, $r_\epsilon(y,z)$ gets closer to the true $r(y,z)$, and the performance of our external study is improved. This indicates that the classifier is well calibrated (since high confidence predictions are more likely to be correct). However, it takes a high value of $\epsilon$ to get a correct approximation of the true association between $y$ and $z$, meaning that we need to discard a large amount of data points from our preliminary study to approximate $r(y,z)$. For example, at $\epsilon=.9$, roughly half of the training instances remain.

\noindent\textbf{Correlation matching method:} For this method, we make the true correlation $r(y,z)$ change between $-0.8$ and $0.8$ and we plot the results on Figure \ref{fig:corr_matching_results}. We observe in the top plot that after adjustment, our estimate $r_{cm}(y,z)$ is within $0.1$ of the true correlation in the worst case against $0.4$ without adjustment. This is a clear improvement in the correlation estimation. (For comparison, achieving a similarly accurate estimate using $\epsilon$ thresholding requires removing 60\% of 1500 instances.) We can also notice that the performance of our preliminary study greatly increases when we improve the estimation of $r(y,z)$, particularly when $r(y,z)$ is high. For example, when $r(y,z)$ is .8, the $F1_z$ improves from ~.77 to ~.9, on average. Thus, correlation matching appears to both recover the true correlation while simultaneously improving the quality of the classifications of $z$.

\begin{figure}[!ht]
  \centering
  \setcounter{subfigure}{0}
  \subfigure[Location/gender dataset\label{fig:fixed_f1z_umbrella_locgender}]{\includegraphics[width=0.45\textwidth]{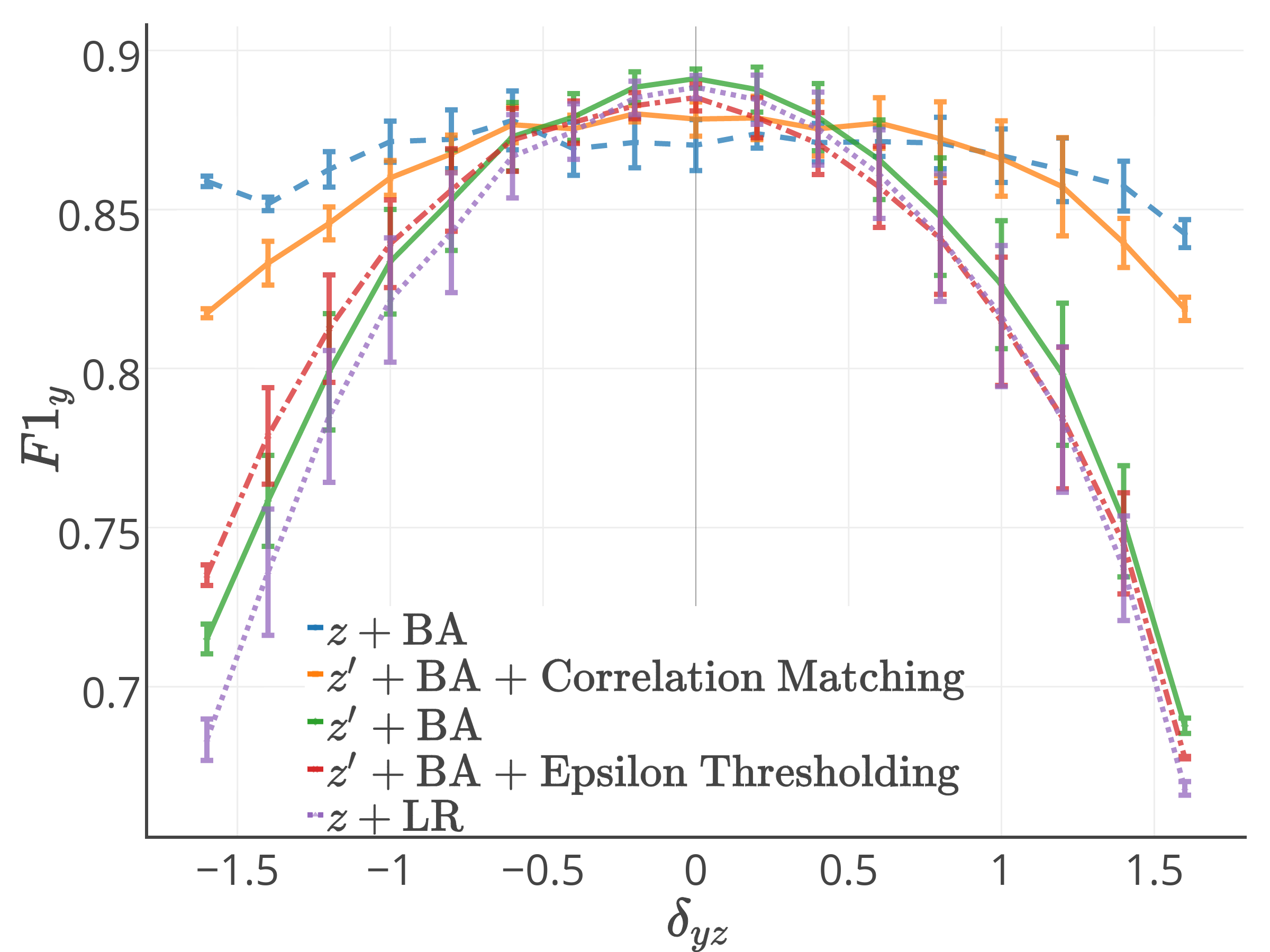}}
  \subfigure[Smoker/gender dataset\label{fig:fixed_f1z_umbrella_smokergender}]{\includegraphics[width=0.45\textwidth]{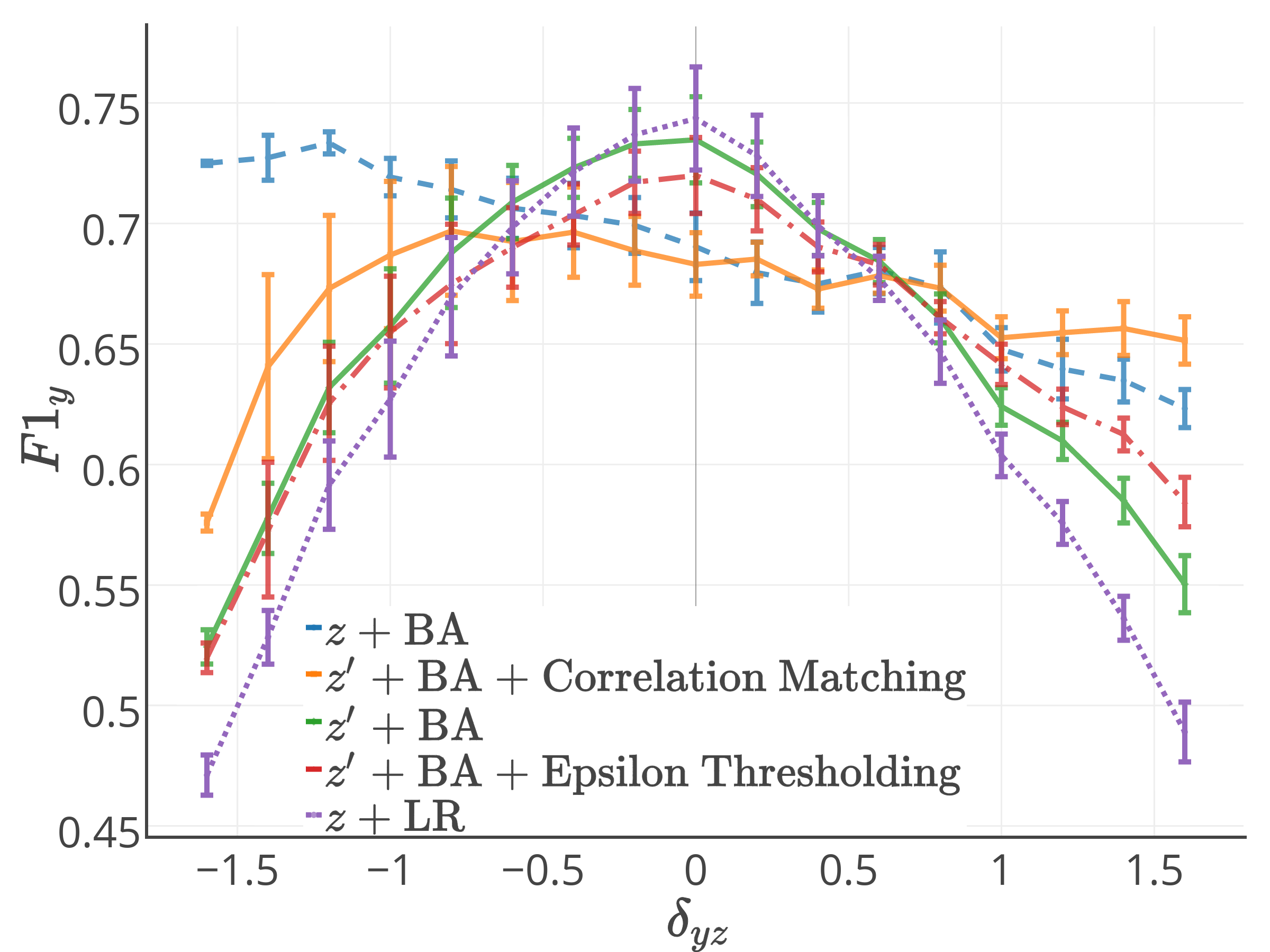}}
  \label{fig:fixed_f1z_umbrella}
  \caption{$F1_y$ of the different adjustment methods when $F1_z$ is fixed to its maximal value vs. logistic regression ($z$ + LR) and back-door adjustment ($z$ + BA) when $z$ is observed.}
\end{figure}

\subsection{Effects of correlation adjustments on $F1_y$}

\begin{figure*}[t]
  \centering
  \setcounter{subfigure}{0}
  \subfigure[No adjustment.\label{fig:vanilla_aba}]{\includegraphics[width=0.32\textwidth]{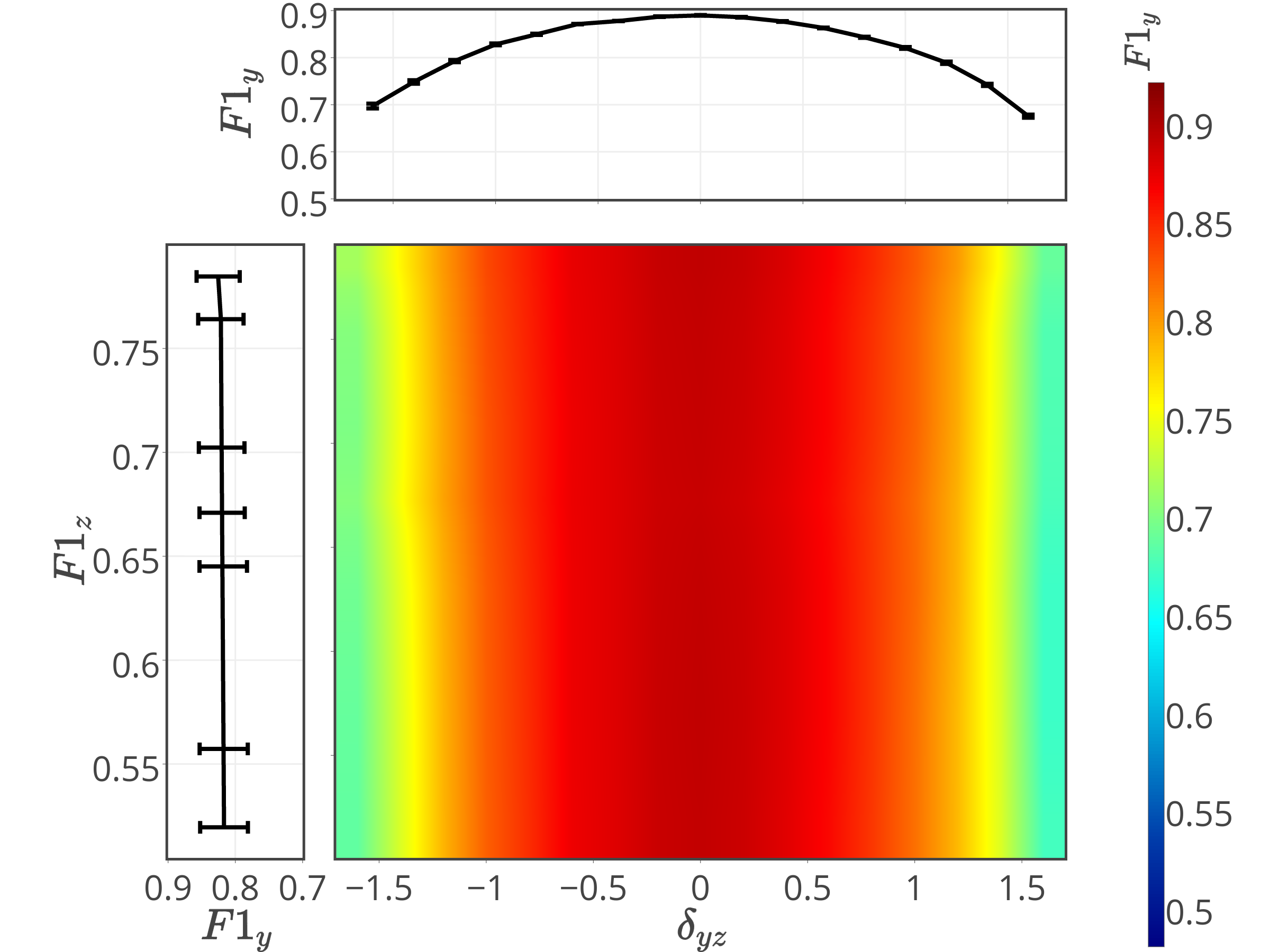}}
  \subfigure[Thresholding at $\epsilon = 0.75$.\label{fig:aba_eps_075}]{\includegraphics[width=0.32\textwidth]{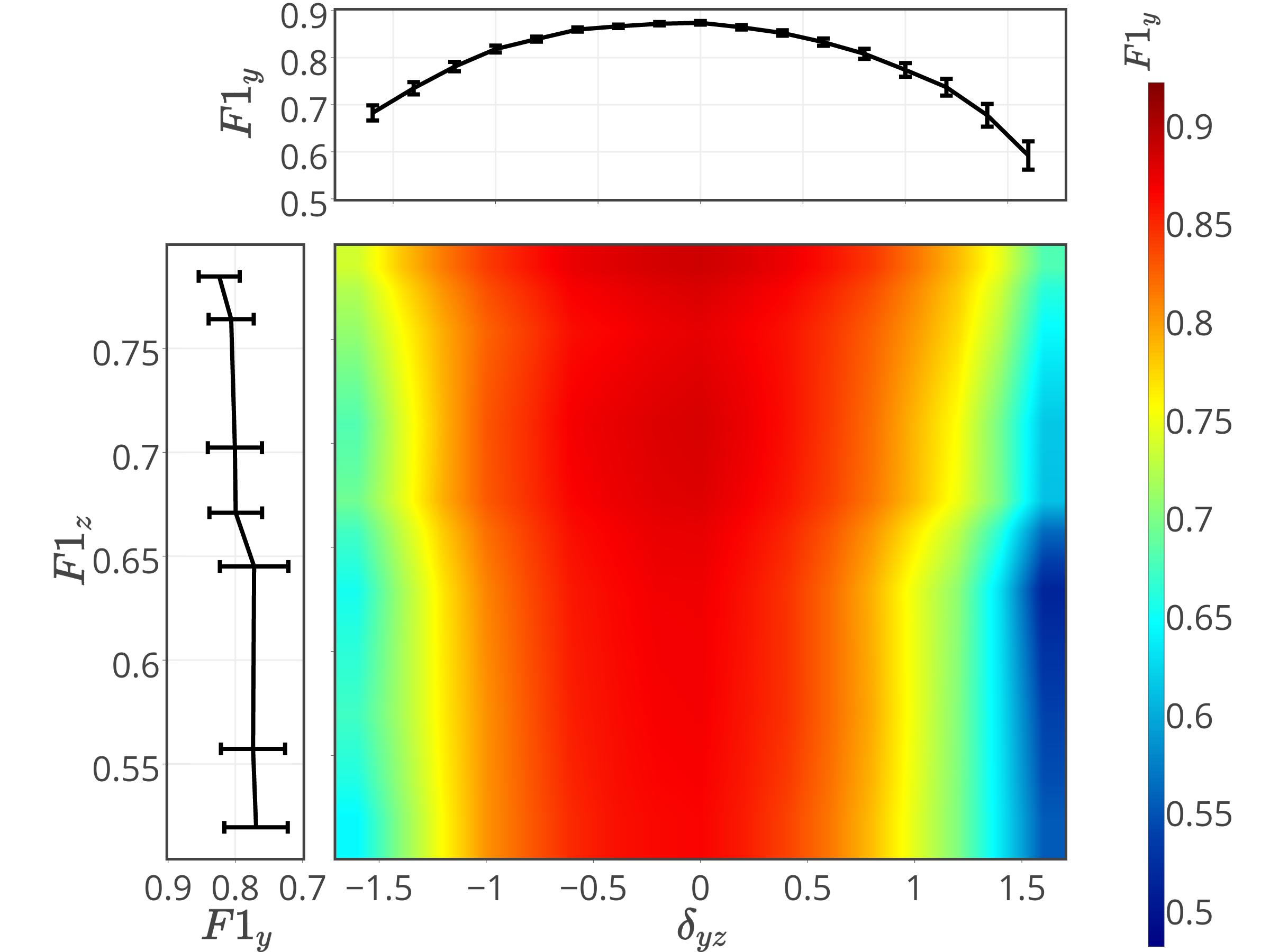}}
  \subfigure[Correlation matching.\label{fig:aba_correlation_1}]{\includegraphics[width=0.32\textwidth]{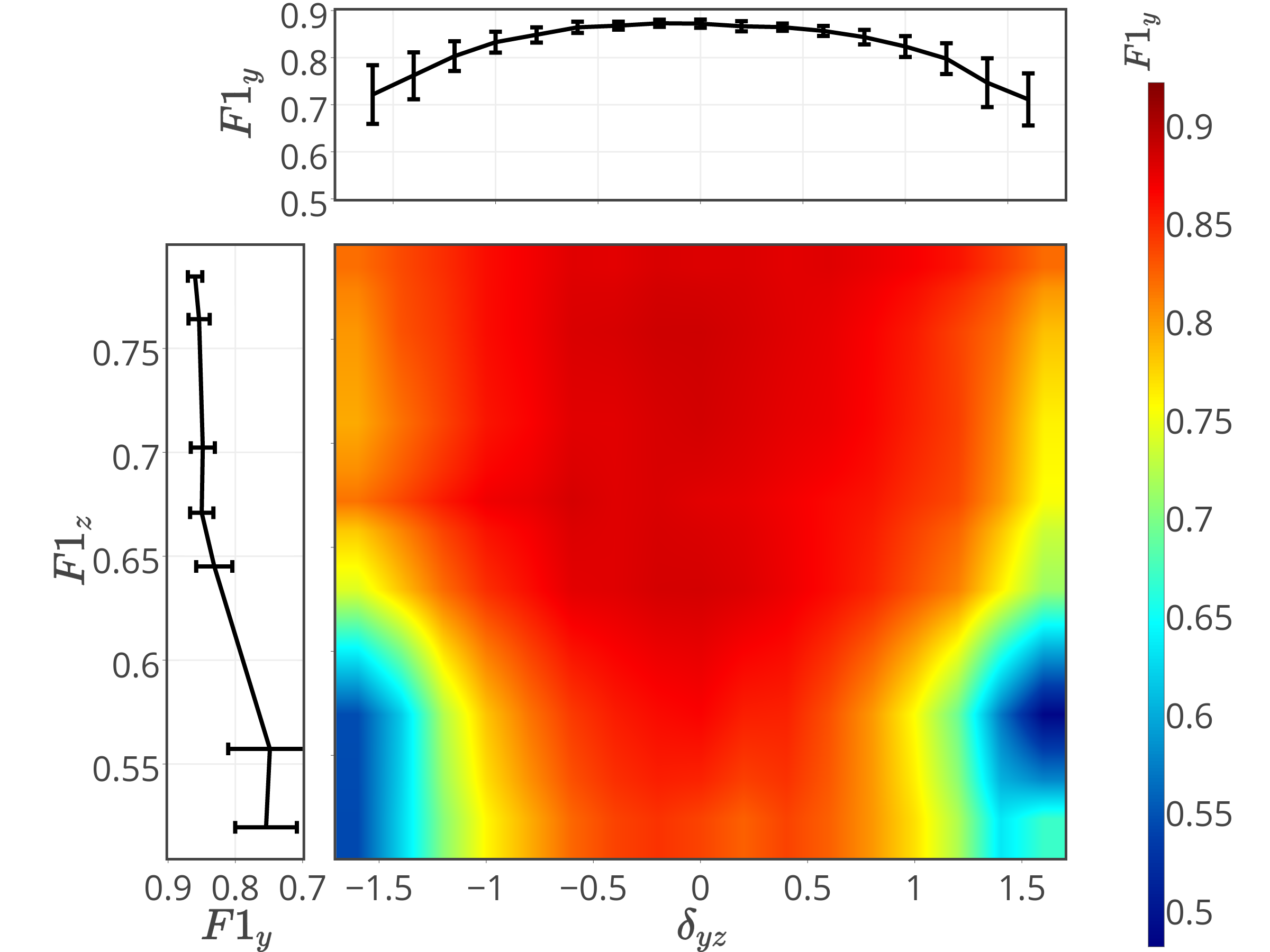}}
  \caption{Experimental results for back-door adjustment with an \textit{unobserved} confounding variable in the location/gender dataset.}
  \label{fig:heatmaps_locgender}
\end{figure*}

\cut{
\begin{figure*}
  \centering
  \setcounter{subfigure}{0}
  \subfigure[No adjustment.\label{fig:vanilla_aba}]{\includegraphics[width=0.33\textwidth]{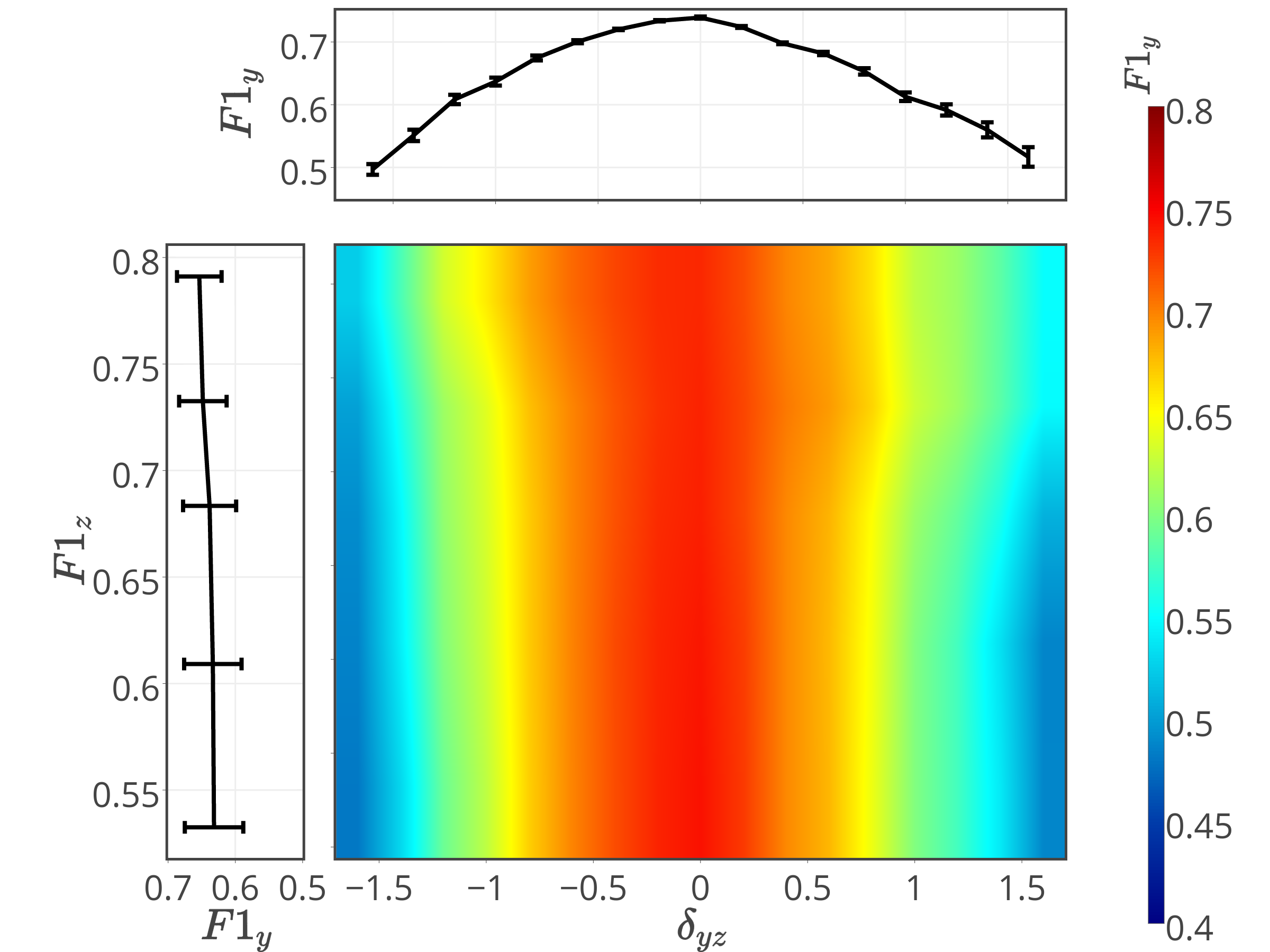}}
  \subfigure[Thresholding at $\epsilon = 0.75$.\label{fig:aba_eps_075}]{\includegraphics[width=0.33\textwidth]{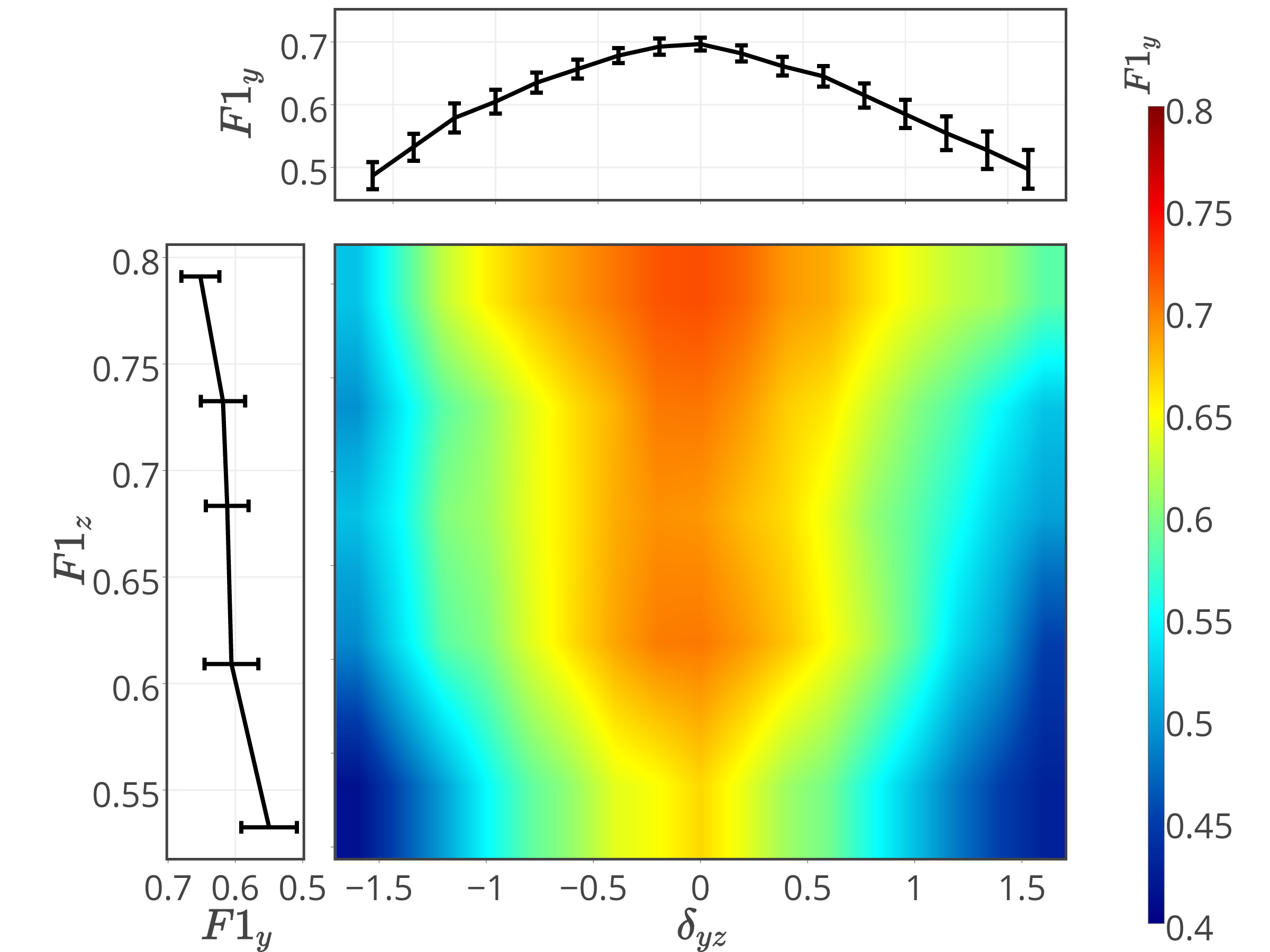}}
  \subfigure[Correlation matching.\label{fig:aba_correlation_1}]{\includegraphics[width=0.33\textwidth]{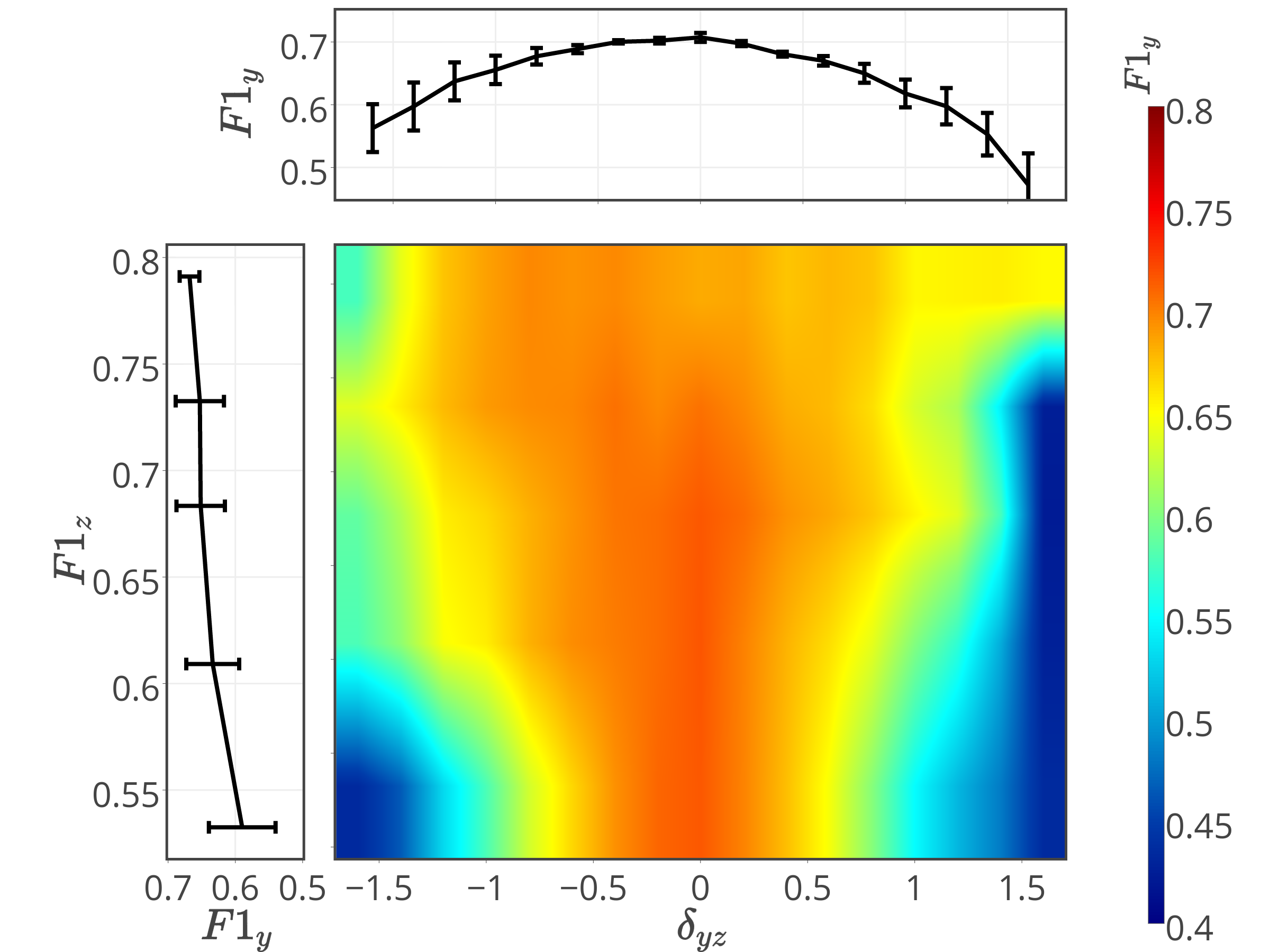}}
  \caption{Experimental results for back-door adjustment with an \textit{unobserved} confounding variable in the smoking status/gender dataset.}
\end{figure*}
}

\begin{table}[!ht]
\centering
\begin{tabular}{lccc}
\toprule
{$\mathbf{F1_z}$} &  {\bf No adjustment} &  {\bf Corr. matching} &  {\bf $\epsilon$ thresh.} \\
\midrule
\textbf{0.784} &       0.0640 &              \underline{0.0212} &              0.0610 \\
\textbf{0.764} &       0.0674 &              \underline{0.0313} &              0.0671 \\
\textbf{0.702} &       0.0677 &              \underline{0.0357} &              0.0803 \\
\textbf{0.670} &       0.0672 &              \underline{0.0345} &              0.0783 \\
\textbf{0.645} &       0.0705 &              \underline{0.0537} &              0.101 \\
\textbf{0.557} &       \underline{0.0715} &              0.124 &              0.0954 \\
\textbf{0.519} &       \underline{0.0709} &              0.0916 &              0.0941 \\
\bottomrule
\end{tabular}
\caption{Standard deviation as a measure of robustness. The smaller the standard deviation, the more robust the model. The most robust model is shown in bold for each $F1_z$ value.}
\label{tab:robustness-table}
\end{table}

\noindent\textbf{Location / Gender}

\noindent\textbf{Fixed $\mathbf{F1_z = 0.784}$:} As our primary result, we report the $F1_y$ obtained by different correlation adjustment methods across a range of shifts in the discrepancy between training and testing. For the Twitter dataset, the best performance we get in the preliminary study is $F1_z = 0.784$. We then compare testing $F1_y$ as $r_{train}(y,z)$ and $r_{test}(y,z)$ vary. The results are shown in Figure \ref{fig:fixed_f1z_umbrella_locgender}. Without any adjustment, the performance we get is close to Logistic Regression. When using $\epsilon$ thresholding, the performance is slightly improved in the extreme cases but only by a few points at most. However, when using the correlation matching method, we improve $F1_y$ by 10 to 15 points in the most extreme cases.  For comparison, the figure also shows the fully observed case ($z + $BA), which uses back-door adjustment on the {\it true} values of $z$. We can see that correlation matching is comparable to the fully observed case, even with a ~20\% error rate on $z$. These results show that by getting a better estimate of the association between $y$ and $z$, we can reduce attenuation bias and improve the robustness of our classifier, even though our observation of $z$ is noisy.

\noindent\textbf{Variable $\mathbf{F1_z}$:} We showed in the previous section that when we use our preliminary study with $F1_z = 0.784$, we can build a robust classifier using the correlation matching method combined with back-door adjustment. We also saw in Figure \ref{fig:broken-ba-vs-lr} that back-door adjustment when $z$ is observed at training time is sensitive to noise in $z$. As a similar study, we want to see how sensitive the correlation adjustment methods are to the quality of $F1_z$. To do so, we increasingly add noise to the dataset used to train the preliminary classifier ($D_z=\{\x_i, \z_i\}$) to make $F1_z$ decrease. \cut{As the $z$ classifier becomes less reliable, we run the same study that we ran in the previous paragraph except that $F_z < 0.784$.} Because we want to visualize $F1_y$ against two variables ($F1_z$ and $\delta_{yz}$), we visualize the results in a heatmap. In order to make the results clear to the reader, here are additional details to understand what is displayed on the heatmap:
The x-axis of a heatmap is $\delta_{yz}$ and the y-axis is $F1_z$. The line plot on the left of the heatmap shows $F1_z$ given $F1_y$ averaged over all possible values for $\delta_{yz}$. The error bars are the standard deviations of $F1_y$, indicating how sensitive the model is to variations of $\delta_{yz}$. \cut{Because we take the standard deviation of $F1_y$ for a given $F1_z$ but across all possible $\delta_{yz}$, a small standard deviation indicates that $F1_y$ is robust to $\delta_{yz}$.} Similarly, the scatter plot above the heatmap shows $F1_y$ given $\delta_{yz}$ averaged over all possible values for $F1_z$. The error bars are the standard deviations of $F1_y$ for the matching $\delta_{yz}$.

Moreover, Table \ref{tab:robustness-table} displays the values of the standard deviations shown in the scatter plot at the left of each heatmap as a measure of robustness. Figure \ref{fig:vanilla_aba} shows the heatmap of results for back-door adjustment when we use the predictions of the preliminary study but none of the methods to fix the mislabeled values in $z'$ are used. Figures \ref{fig:aba_eps_075} and \ref{fig:aba_correlation_1} respectively show the heatmaps of results when we use $\epsilon$ thresholding with $\epsilon=0.75$ and correlation matching.
Similar to Figure~\ref{fig:fixed_f1z_umbrella_locgender}, $\epsilon$ thresholding only brings small improvement to no adjustment at all.
Furthermore, when $F1_z$ decreases, the correlation adjustment using $\epsilon$ thresholding is performing worse than when we are not doing any correlation adjustment as well as it is less robust. Clearly, the $\epsilon$ thresholding method is more sensitive to the quality of the preliminary study than the other methods.

\cut{Furthermore, it looks like $F1_z$ does not affect the $y$ classifier in the first two heatmaps. This may be explained by the fact that when $F1_z = 0.784$ (our maximum value for $F1_z$), there are so many mislabeled points in $z$ that no adjustment or $\epsilon$ thresholding methods perform as well as logistic regression. Because the underlying model of back-door adjustment is logistic regression, its performance has a lower bound which is performance of logistic regression. Here, we lack a better $z$ classifier (with $F1_z > 0.784$) that would allow $\epsilon$ thresholding to perform better.}

The correlation matching method (Figure~\ref{fig:corr_matching_results}) does outperform the other methods in robustness and $F1_y$ for most of the cases but when $F1_z < 0.645$, as we can see by the wider range of red values in Figure~\ref{fig:aba_correlation_1}. In this latter case, it performs worse than the method without adjustment. This method is also sensitive to the quality of the preliminary study as we can see that the averaged $F1_y$ decreases with $F1_z$. \cut{However, it can be sensitive to the quality of the preliminary study, as $F1_y$ drops to less than $0.7$ in the extreme cases when $F1_z \leq 0.74$. We believe this is partly attributed to the poor estimates $\hat{r}$ produced by such an inaccurate classifier. While addressing this sensitivity is an open issue, we note that in most practical settings, we would expect a binary classifier for confounding variables to be greater than 70\%. Furthermore, these graphs are limited by the maximum accuracy of the preliminary classifier in this data ($F1_z=0.784$). In other domains, accuracy may be much higher, increasing the range in which correlation matching is most effective.}
Let us remind one more time that we are considering here only preliminary studies with an $F1_z$ of at most $0.784$. Therefore, $F1_z$ could be up to 22 points greater with a different dataset. This would hopefully lead to similar results than when $F1_z = 0.784$ with correlation matching and better results in $F1_y$ and robustness with $\epsilon$ thresholding.

\noindent\textbf{Smoker / Gender}

\noindent\textbf{Fixed $\mathbf{F1_z = 0.791}$:} Similarly to the previous experiment, we report $F1_y$ while making $\delta_{yz}$ vary as our primary result in Figure~\ref{fig:fixed_f1z_umbrella_smokergender}. We observe that predicting if a user smokes or not is a much more difficult task than our previous binary location prediction task, as the maximum yielded $F1_y$ is around .75 when it was approximately .9 in the previous task. We also notice that the robustness of the back-door adjustment methods is not as good as for the location/gender dataset. The correlation matching method manages to performs closely to $z + \text{BA}$ for $\delta_{yz} \geq -0.75$ and outperforms all other methods for $\delta_{yz} \geq 1$ but we also witness an accuracy drop on the left part of the plot. In addition to this drop, our two most robust methods ($z + \text{BA}$ and correlation matching) are outperformed by approximately 5 points when there is no difference between the training correlation and the testing correlation (when $\delta_{yz} = 0$). 

\noindent\textbf{Variable $\mathbf{F1_z}$:} When making $F1_z$ vary with the smoker/gender dataset, we observe comparable outcomes as the ones displayed in the heatmaps of Figure~\ref{fig:heatmaps_locgender} but with a lesser overall accuracy. As back-door adjustment was not performing as well as with the location/gender dataset in the fixed $F1_z$ case, it logically also does not perform as well when $F1_z$ varies. If we obtain a V-shaped heatmaps similar to Figures~\ref{fig:aba_eps_075} and \ref{fig:aba_correlation_1}, the slope indicating that the classifier's' performance deteriorates when $F1_z$ decreases is steeper. This may show that our adjustments methods are more sensitive to noise in the confounding variable when the classification task is overall harder. We do not display the resulting heatmap for the smoker/gender experiment in this paper for brevity but we will make the dataset and the code to reproduce the results available online.

\section{Conclusion}

In this paper, we have proposed two methods of using back-door adjustment to control for an unobserved confounder. Using two real-life datasets extracted from Twitter, we have found that correlation matching on the predicted confounder associated with back-door adjustment can retrieve the underlying correlation $r(y,z)$ and perform closely to back-door adjustment with an observed confounder. We also showed that $\epsilon$ thresholding can be used to slightly improve the predictions compared to logistic regression. If $\epsilon$ thresholding will not be able to adjust for the unobserved confounder $z$ when $F1_z < 0.75$, we showed that correlation matching provides a way to adjust for an unobserved confounder and outperform plain back-door adjustment as long as $F1_z > 0.65$. In future work, we will consider hybrid methods that combine $\epsilon$ thresholding and correlation matching to increase robustness as $F1_z$ decreases.

\cut{We also saw that correlation matching drops dramatically when $F1_z$ decreases, our future work will focus on fixing this problem by making sure that the correlation estimate we are optimizing for is correct. We will also study the effect of combining correlation matching with $\epsilon$ thresholding in order to get the best out of the two methods.}

\section*{Acknowledgments}
This research was funded in part by the National Science Foundation under
awards \#IIS-1526674 and \#IIS-1618244.

\bibliographystyle{aaai}
\bibliography{icwsm17}{}

\end{document}